# Impact of Narrow Lanes on Arterial Road Vehicle Crashes:

# A Machine Learning Approach

**Mohammed Elhenawy**

Center for Sustainable Mobility, Virginia Tech Transportation Institute

3500 Transportation Research Plaza, Blacksburg, VA 24061

E-mail: mohame1@vt.edu

Phone: (540) 231-0278

**Arash Jahangiri**

Lecturer, San Diego State University

5500 Campanile Dr, San Diego, CA 92182

E-mail: AJahangiri@mail.sdsu.edu

Phone: (540) 200-7561

**Hesham Rakha (corresponding author)**

Samuel Reynolds Pritchard Professor of Engineering, Dept. of Civil & Environmental Engineering

Courtesy Professor, Bradley Dept. of Electrical and Computer Engineering

Director, Center for Sustainable Mobility

Virginia Tech Transportation Institute

3500 Transportation Research Plaza (0536), Blacksburg, VA 24061

E-mail: HRakha@vtti.vt.edu

Phone: (540) 231-1505

Fax: (540)-231-1555

**Abstract**

In this paper we adopted state-of-the-art machine learning algorithms, namely: random forest (RF) and least squares boosting, to model crash data and identify the optimum model to study the impact of narrow lanes on the safety of arterial roads. Using a ten-year crash dataset in four cities in Nebraska, two machine learning models were assessed based on the prediction error. The RF model was identified as the best model. The RF was used to compute the importance of the lane width predictors in our regression model based on two different measures. Subsequently, the RF model was used to simulate the crash rate for different lane widths. The Kruskal-Wallis test, was then conducted to determine if simulated values from the four lane width groups have equal means. The test null hypothesis of equal means for simulated values from the four lane width groups was rejected. Consequently, it was concluded that the crash rates from at least one lane width group was statistically different from the others. Finally, the results from the pairwise comparisons using the Tukey and Kramer test showed that the changes in crash rates between any two lane width conditions were statistically significant.

Keyword; Safety, Crash rate, lane width, Machine Learning, Random forest, Least Squares Boosting.

1. **Introduction**

The number of operating vehicles in the world will, at least, double by 2050 [1]. This rapid growth in the number of vehicles increases motor vehicle crashes. Motor vehicle crashes result in significant economic and societal losses in the USA. The traffic safety facts prepared by the national highway traffic safety administration (NHTSA) showed that in 2014, there were 32,675 people who died in motor vehicle crashes in the US. Moreover there were 2.3 million people injured in crashes in the same year [2]. The last report published in January 2016 which shows the crash statistics for the first nine months of 2015 indicates an increase of approximately 9.3 percent in fatalities compared to the same period of 2014 [3]. A recent report released by the U.S. Department of Transportation's National Highway Traffic Safety Administration (NHTSA) shows the high economic cost of the motor vehicle crashes in the United States. The report indicates that $836 billion in economic loss and societal harm which is almost $800 for each person living in the United States based on calendar year 2010 data [4].

Because of the economic and social losses caused by motor vehicle crashes, traffic safety is an active research area that has many research sub-areas. Crash data analysis is the most prominent sub-area of traffic safety that has the goal of identifying the factors that lead to crashes and then developing countermeasures that reduce crash frequency. This kind of research requires detailed driving data such as that recorded in the vehicle black boxes. However, detailed data are not always available so that most of the current research studies focus on analyzing the crash frequency datasets occurring in some geographical space [5]. Crash frequency research study's the factors that can be used to explain the variation in the crash frequency and hence identify significant factors. By studying the significant factors that lead to crashes, accidents, injuries, financial losses, delays, the impact of these crashes on the transportation system can be reduced. The recent advances in technology that enable continuous collecting and recording of traffic data and cheap storage has created a new trend in traffic safety research. The new research uses the traffic conditions immediately before crashes occur to recognize traffic patterns that are common before crashes. This new research does not require aggregating the crash data because of the availability of the traffic conditions before each crash [6].

## 2. Literature review

Modeling of crash data using road geometric design features including the number of lanes and traffic characteristics such as the Average Annual Daily Traffic (AADT), is an area of significant research interest. In general, crash data can be in the form of crash frequency or crash rate data. Crash frequency is aggregated count data and needs special regression methods that are developed to model non-negative integers. Crash rate is derived from crash frequency by normalizing the crash frequency based on certain measures of exposure. There are two main approaches to model and predict crash data, statistical regression models and machine learning models. Statistical regression models are parametric techniques that require assumptions about the distribution of crash data and a well-defined mathematical model that relates crash data to the model predictors.

There are many statistical tools used to analyze crash data. An early study used multiple linear and Poisson regression to build a model that linked the large truck crash and traffic characteristics and road geometry [7]. It should be noted that multilinear regression assumes a normal error structure and constant error variance, which is typically not satisfied in crash data. Moreover, Poisson regression assumes the mean and variance of the data are the same, which is typically not the case in crash data. Usually, crash data has a variance that exceeds its mean which is known as over-dispersion [8]. In some rare cases, when the sample mean is very low, the crash data suffers from under-dispersion. Modeling under-dispersed data using traditional count-data may result in incorrect parameter estimates [5].

In a study, Quasi-Poisson regression was used to analyze and link the response, which is a combined measure of the crash frequency and corresponding harmfulness, with several key explanatory variables such as pavement condition [9]. Quasi-Poisson was proposed to deal with the over-dispersion problem. The quasi-Poisson model has the mean of $\lambda_i$ and variance of $\varphi\lambda_i$, and also Quasi-Poisson leaves the dispersion parameter $\varphi$ unfixed at 1 and estimates this new parameter from the data. The quasi-model formulation gives the same coefficient estimates as the original Poisson model, but the inference is modified to consider the over-dispersion [9]. Negative binomial regression is another model used to analyze crash frequency in many publications [10-12]. In the negative binomial model, the distribution mean equals $\lambda$ and variance equals $\lambda + k\lambda^2$ where $\lambda > 0$ and $k > 0$. In this model, the dispersion parameter $(1 + k\lambda)$ depends on the mean and is not constant as in the quasi-Poisson. Quasi-Poisson linearly relates the variance of the count data to the mean, whereas for the negative binomial the variance is modeled as a quadratic function of the mean.

Real crash data sets typically have many observed zeros than assumed by the count model such as the Poisson distribution and the negative binomial distribution. This phenomenon is called "zero-inflated". Zero-inflated models are proposed to deal with the zero-inflated phenomena [13]. Zero-inflated models are two-component mixture models; the first component is a point mass at zero while the other component is a count distribution such as Poisson or negative binomial. Due to the ability of the zero –inflated models to handle datasets that have a large number of zero-crash observations, it has been used in crash data analysis [14-16]. The Hurdle model is another tool used to model crash frequency data [17]. It is a mixture model designed to deal with zero-inflated count data. There is an important distinction between the Zero-inflated and hurdle models in how they interpret and analyze zero counts. A zero-inflated model assumes that zeros are generated from both the point mass and the count distribution component whereas, the hurdle model assumes the zeros are produced only from the mass point component. A hurdle model assumes the positive count data comes from a truncated count distribution such as a truncated Poisson or a truncated negative binomial distribution. Zero-inflated and hurdle models can yield different results with very different interpretations. In general, over-dispersion can happen because of two sources: heterogeneity of the population and excess of zeroes .The heterogeneity is observed when the population has natural groups (clusters) and can be divided into many homogeneous groups. Recently researchers started applying finite mixture models to the analysis of crash data [18, 19].

Due to the assumptions associated with statistical regression models and advances in machine learning techniques, researchers have started to use machine learning algorithms to model crash data [20, 21]. Machine learning techniques such as neural networks have the disadvantage that it is a black box, and usually we cannot identify the relationships between crash frequency and input predictors. Recently, the genetic programming (GP) model was used for real-time crash prediction. GP models have the advantage that it is not a black box [22].

### 3. Machine Learning Methods

#### 3.1. Random Forest (RF)

The random forest (RF) method, proposed in 2001 [23], creates an ensemble of decision trees. For each tree, a subset of features is randomly selected to grow the tree. The building of decision trees uses a greedy and recursive algorithm that starts from a root where the entire data are in one node. Subsequently, a tree is grown by splitting the data using a binary splitting approach. The chunk of the data at the parent node is proportioned between the resultant leaves to minimize the objective function. In order to predict the response for a new unseen test observation, it is pushed down by going through the tree from the root to a leaf. The final leaf determines the response of the test observation [24]. While growing the tree, the data is divided by employing a criterion in several steps or nodes. In practice, the mean-square error (MSE) of the responses is used for regression. The response of a test observation is obtained by averaging the predictions from all the individual regression trees on this unseen test observation.

the Random Forest technique offers several advantages, namely: this approach is computationally simple and quick to fit, even for large problems; it handles categorical variables naturally and thus there is no need to create extra dummy variables; it can be applied to both regression and classification problems; it can handle highly non-linear variable interactions; it automatically conducts variable selection by ranking them based on each variables' individual contributions; and the over-fitting issue does not occur (might be an issue in only rare cases). In addition, tuning parameters can be easily optimized. Furthermore, the RF produces an unbiased estimate of the generalization error when the trees are growing. This measure called Out-of-Bag (OOB) error, is similar to the cross validation procedure and can be used for model selection and validation. In this paper, however, the OOB error was not applied because we were interested in converting the crash rates back to crash counts and since the OOB error is computed internally it was not possible to do the conversion. Therefore, a test set was used for model selection and validation. The random forest has just few limitations: the observations need to be independent and thus the data aggregation was carried out in this paper to resolve this issue. Also, the final estimation (or prediction) result is obtained through averaging many tree models; so model interpretation becomes more difficult compared to say a single tree model.

### 3.2. Least squares boosting (LSBoost)

In machine learning, boosting is one of the state-of-the-art learning concepts. The first boosting algorithm was originally proposed to build classifiers. Subsequently, the boosting concept was extended to build regression models. The main idea of boosting is combining the outputs of many "weak" classifiers/regression models to create a very powerful "committee."

A weak learner can vary from a very basic classifier, such as a two terminal node classification regression tree to the powerful support vector machine (SVM). Boosting algorithms greedily apply a sequence of weak learner algorithms to repeatedly modify versions of the training data to build a learner community $L_k(x), k = 1, 2, \ldots, K$.

Least squares boosting (LSBoost) fits regression models by minimizing the mean-squared error objective function. At each iteration $M$, the LSboost fits a new regression model to the difference between the true response and the sum of the prediction of all the $M - 1$ regression models fitted previously.

### 4. Problem Definition and Formulation

This paper builds two crash prediction models using two state-of-the-art machine learning algorithms. The built models use different explanatory variables (predictors) including the "lane width" variable. The two models were built using 10 years-worth of crash data in four cities in Nebraska. The better of the two models was used to quantify the impact of narrow lanes on the safety of arterial roads. The variables used in this study are presented in Table 1. To pursue this goal, the Random Forest (RF) algorithm, and the LSBoost algorithm are adopted to develop the models.

Table 1 Variable Definition

| Variable | Definition | Variable | Definition |
|---|---|---|---|
| $X_1$* | Crash Rate* | | |
| $X_2$ | Crash Counts | $X_9$ | One-way Indicator |
| $X_3$ | Section Length | $X_{10}$ | Number of Lanes |
| $X_4$ | Section Number | $X_{11}$ | Road Classification Indicator |
| $X_5$ | Year | $X_{12}$ | Median Indicator |
| $X_6$ | Shoulder Indicator | $X_{13}$ | Lane Width |
| $X_7$ | Speed Limit | $X_{14}$ | Central Business District Indicator |
| $X_8$ | On-street Parking Indicator | $X_{15}$ | AADT per lane |

* The variable "crash rate" denoted by $X_1$, which is not in this table, will be defined later in the "model development" subsection.

The RF regression model is formulated as

$$RF(V, \{\theta_k\}) = \{t(V, \theta_1), t(V, \theta_2), \dots, t(V, \theta_K)\} = \left(\frac{1}{K}\right) \sum_{k=1}^{K} t(V, \theta_k) \qquad \text{Equation 1}$$

Here $V$ represents the input vector including different variables (e.g. $X_2, X_3, \dots$), $\theta_k$ denotes the randomization vector for data subsampling and variable selection, $K$ is the number of trees, and $t(V, \theta_k)$ is the $k^{th}$ regression model obtained from a single tree. As was presented in the formulation, the model consists of an ensemble of $K$ regression trees (i.e. $t(V, \theta_1)$) and the final result is obtained through averaging over the results of all $K$ trees.

The LS AdaBoost regression model can be formulated as

$$LSboost(V, \{\theta_k\}) = \{t_1(V, \theta_1), t_2(V, \theta_2), \dots t_n(V, \theta_n), \dots, t_k(V, \theta_K)\} \qquad \text{Equation 2}$$

$$= \sum_{k=1}^{K} t_k(V, \theta_k)$$

Where $V$ represents the input vector including different variables (e.g. $X_2, X_3, \dots$), $\theta_k$ defines the split variables and split points at the internal nodes, and the predictions at the terminal nodes, $K$ is the number of trees, and $t_n(V, \theta_k)$ is the $n^{th}$ regression tree which models difference between the true response and $\sum_{k=1}^{n-1} t_k(V, \theta_k)$.

5. **Data Analysis**

The following subsection will present the analysis results of the data. In implementing the RF technique, the R software [25] and the package "RandomForest" [26] were used. The LSBoost was implemented in Matlab. In addition, the comparison between distributions was made using the package "GMD" [27].

### 5.1. Model Development and Justification

When using RF and LSBoost, the individual observations need to be independent. However, the data for the present study included observations that were not completely independent because each road section had several observations from different years. Consequently, observations from different years for each section were averaged and aggregated into single observations leading to a total of 1,818 observations (after removing missing and duplicate data). The data aggregation had another benefit with regards to the nature of traffic crashes, which are considered as rare and random events. The data aggregation provided more exposure and thus more realistic crash counts for each section and reduced the number of zero counts. Furthermore, 80% of the data was used for model development (i.e. training set), and the remaining 20% was used for model validation (i.e. testing set). Moreover, instead of using the actual crash counts as the dependent variable, crash rates based on section length and traffic volume were used as widely applied in the literature [28, 29]. Equation 3 presents the adopted dependent variable (response).

$$X_1 = \frac{X_2}{X_3 \cdot (X_{10} \cdot X_{15})^p} * \frac{10^6}{365^p} \qquad \text{Equation 3}$$

where $X_1$ is the crash rate (crash counts per million vehicle mile traveled per day), $X_2$ is the yearly crash counts, $X_3$ is the section length in miles, $X_{15}$ is the section AADT (veh/day/lane), $X_{10}$ is the number of lanes, and $p$ is the exposure measure, which is assumed to be 0.8 as recommended in [28].

In order to compare the two adopted machine learning algorithms, the number of trees used by each model needs to be determined to achieve the best possible results. Sensitivity analysis was carried out to find the best model parameters as shown in Figure 1 and Figure 2. After about 100 trees, the error rates is stabilized; thus the number of trees in both models was selected as 200 trees.

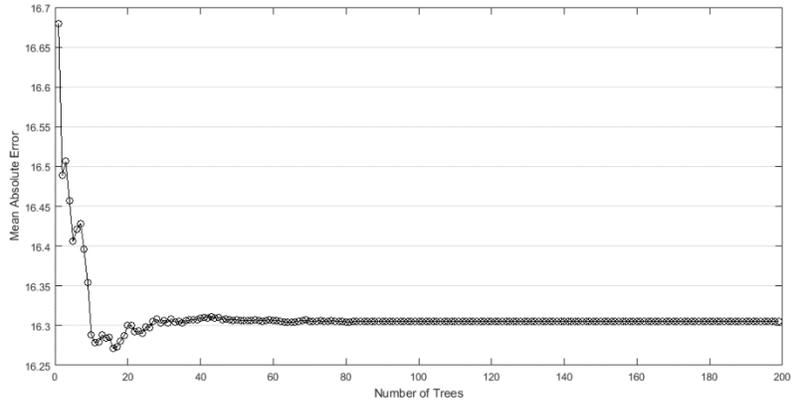

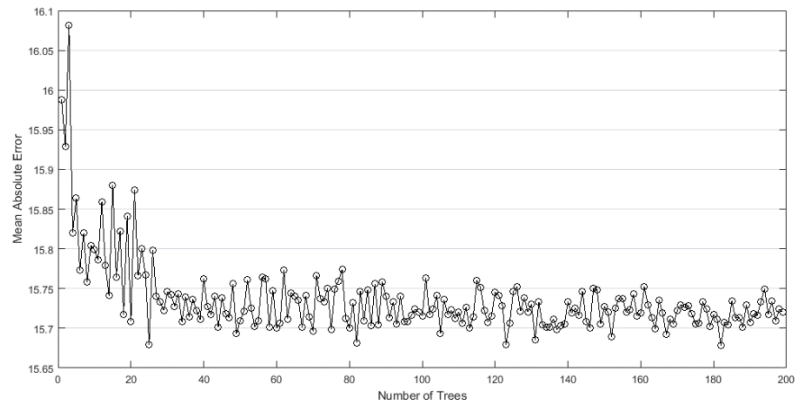

Figure 1 the Top Panel Shows the MAE of LSBoost Regression Model and Bottom Panel Shows the MAE of Random Forest Regression Model

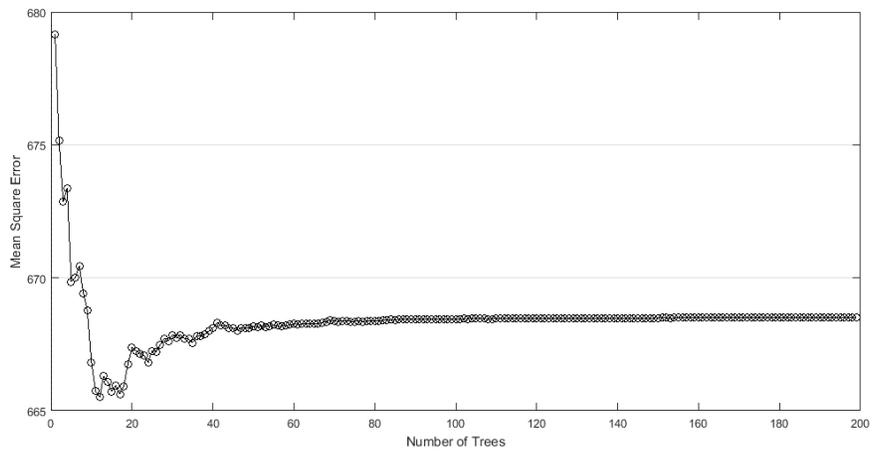

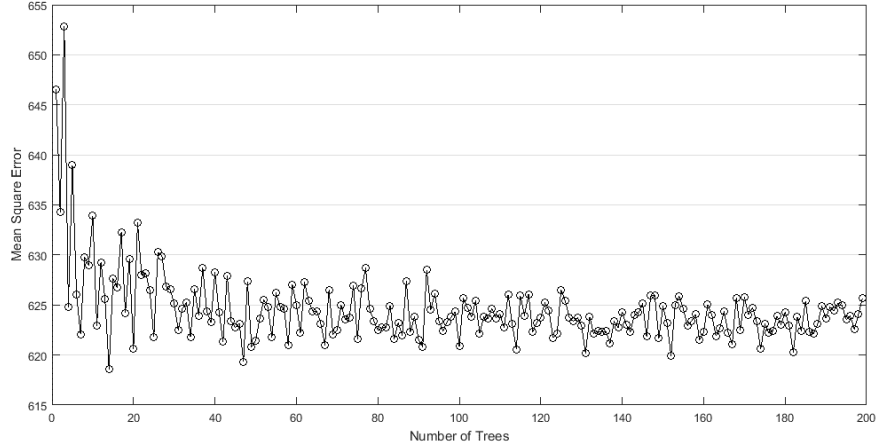

Figure 2 the Top Panel Shows the MSE of LSBoost Regression Model and Bottom Panel Shows the MSE of Random Forest Regression Model

**5.2. Model Adequacy Check**

The mean absolute error (MAE) and mean squared error (MSE) resulting from the final models (i.e. using 200 trees) are presented in Table 2. Moreover the histogram intersection, a measure of similarity between two histograms, was used to quantify the similarity of the histogram generated from the model was to the ground truth crash rate. The higher the value of the histogram intersection the better the model.

$$H(y) \cap H(\hat{y}) = \frac{1}{Q} \sum_{q=1}^{Q} \min(H_q(y), H_q(\hat{y}))  \qquad \text{Equation 4}$$

Here $H(y)$ and $H(\hat{y})$ are the histograms of the ground truth and predicted crash rates, respectively. According to the different measures in this table, which are all based on crash rates, the RF model is better than the LSBoost model. Furthermore, the RF model contributed to a low error rate and also it was able to produce a very similar distribution compared to the actual crash rate as reflected by the histogram similarity measure. Hence, the RF model was used to evaluate the impact of the lane width on the road safety.

Table 2 LSBoost and Random Forest Regression Models: Adequacy Check

|         | Mean Absolute Error | Mean Squared Error | GM Distance |
|---------|---------------------|--------------------|-------------|
| **LSBoost** | 16.31               | 669.42             | 8.96        |
| **RF**      | 15.73               | 617.02             | 10.07       |

### 5.3. Quantify the Impact of the Lane Width on the Road Safety

We studied the importance of the different predictors used in the RF model and Figure 3 shows the variable importance plot based on two measures; the first is the percent increase in MSE if a variable is dropped and the second is the percent increase in node purity when a variable is dropped. According to this figure, the "lane width" variable (i.e. $X_{13}$) ranked the second based on the first measure and the fourth based on the second measure, which shows that this variable is among the most important variables in predicting the number of crashes. Since the RF model represents the relationship between the crash rate and the explanatory predictors including the lane width, we used the model to generate (simulate) a new balanced data set. In order to generate the new dataset, the values of all predictors in each observation were kept unchanged except the integer value of lane width which was varied from 9 to 12 ft. For each new observation, the response was computed using the RF model. The size of the new dataset was four times the size of the original dataset. Figure 4 presents a box plot and a marginal plot for the "lane width" variable using the new dataset that shows how the crash counts change by changing the lane width. The predicted values in the marginal plot were the averages over all predicted values for each lane width category (i.e. the associated box plot).

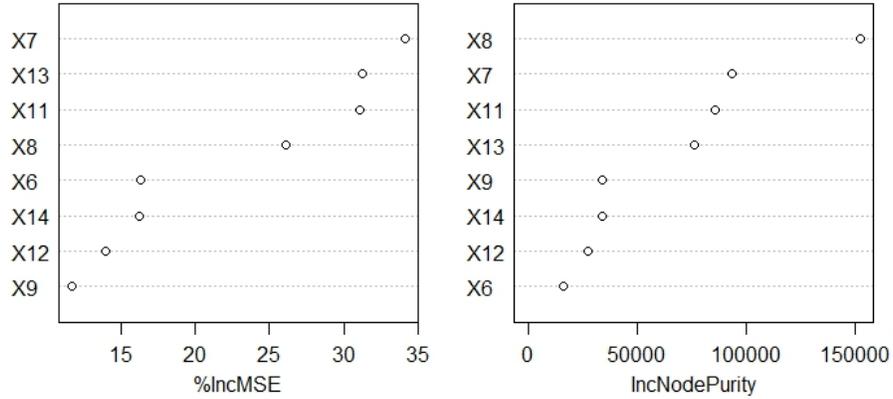

Figure 3 Variable Importance Ranking

According to the plots presented in Figure 4, narrower lane widths (i.e. 9 and 10 ft wide) are associated with higher crash rates compared to the wider lane widths. For example, on average, there was approximately a 25% increase in the crash rate when going from the lane width of 12 to 10 feet.

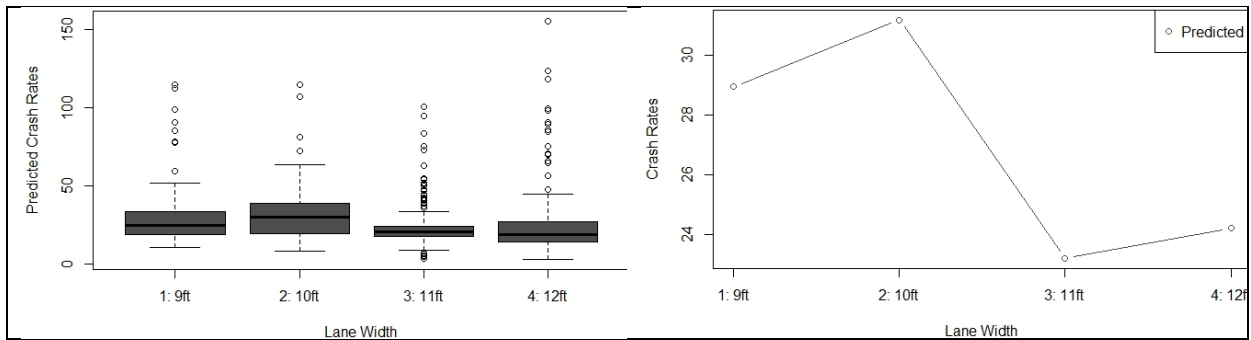

Figure 4 Effects of Lane width on Crash Rates

Changes in crash rates are presented in

Table 4. The percent changes were calculated based on the total average rate of 26.73. In order to verify whether these changes were statistically significant, the Analysis of Variance (ANOVA) test was considered first. However, since the data did not meet the assumptions required to conduct ANOVA (e.g. normality, variance equality), a non-parametric test, namely the Kruskal-Wallis test, was conducted. Table 3 shows the test results. According to the Kruskal-Wallis test, based on the very small p-value, the null hypothesis of equal means for predicted values from the four lane width groups was rejected. Consequently, it was concluded that the crash rates from at least one lane width group is statistically different from the others.

Table 3 Kruskal-Wallis Test Results

| Chi-Squared | Degree of Freedom | p-value |
|---|---|---|
| 786.9032 | 3 | 2.2e-16 |

Table 4 Percent Changes in Crash Rates

| Lane Width (ft) | Lane Width (ft) | Percent Change in Crash Rate |
|---|---|---|
| 10 | 9 | -7.7 |
| 11 | 9 | +20.7 |
| 12 | 9 | +17.3 |
| 11 | 10 | +28.4 |
| 12 | 10 | +25 |
| 12 | 11 | -3.3 |

Subsequently, pairwise comparisons using the Tukey and Kramer (Nemenyi) test was conducted. The results are presented in Table 5. This table shows the p-values corresponding to the pairwise comparisons. Small p-values show that the changes in crash rates between any two lane widths are statistically significant.

Table 5 Pairwise comparisons using Tukey and Kramer (Nemenyi) test: Crash Rates as Dependent Variable

| Lane width | 1: 9ft | 2: 10ft | 3: 11ft |
|---|---|---|---|
| **2: 10ft** | 1.2e-08 | - | - |
| **3: 11ft** | < 2e-16 | < 2e-16 | - |
| **4: 12ft** | < 2e-16 | < 2e-16 | 0.0029 |

## 5.4. Crash Rate versus Crash Count as Dependent Variable

The selection of the dependent variable is critical in any analysis. In this paper, the crash rate was used to be consistent with the literature. However, one may consider the actual crash counts as the dependent variable. To have a comparison, the analysis was repeated with the actual crash counts as the dependent variable and the variables "section length", "number of lanes", and "AADT" were included as additional independent variables. Similar patterns were found regarding the impacts of the "lane width" variable as shown in Figure 5. However, the percent changes in prediction means were not statistically significant in all pairwise comparisons. For example, the predicted values from the lane width of 9 and 10 feet resulted in a large p-value (i.e. 0.9971) and so the null hypothesis was not rejected which means the mean difference was not statistically significant. Similarly, the mean difference of the 11-feet and 12-feet lane width was not statistically significant (i.e. p-value of 0.9025). However, considering the lane width pairs of (9, 11), (10, 11), (9, 12), and (10, 12), the mean differences of the predicted crash counts were statistically different in each case. Using either the crash rates or the actual counts, the narrower lane widths (9 and 10 ft) increase the probability of crashes (or crash rates) compared to wider lane widths (11 and 12 ft). Thereby, it is important to assess different independent variables to have a better understanding of the variable impacts.

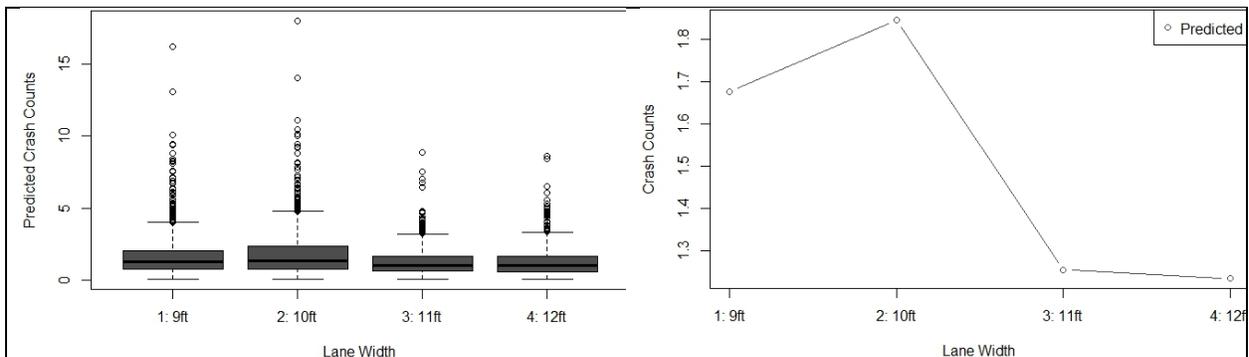

Figure 5 Effects of Lane width on Crash Counts

### 5.5. Data Limitations

Several factors can lead to traffic crashes and thus it would be beneficial to have more detailed information in the data to lower the error rates. For example, a crash may occur due to the driver distraction or aggressiveness and in this particular example the lane width may not be a contributing factor. Therefore, without having detailed information the inference may be misleading. The variables that can be added to improve the model performance include: (1) driver factors; distraction, aggressiveness, drowsiness, (2) road factors; pavement conditions, marking conditions, (3) vehicle factors; vehicle actual speed, vehicle problems, and (4) environment factors; snow, icy road, fog.

### 6. Conclusion

In this paper we adopted two machine learning algorithms to analyze crash data and evaluate the impact of lane width on the safety of arterial roads. The machine learning techniques applied in this study are non-parametric approaches and no formal distributional assumption was needed. The RF model showed a better performance compared to the LSBoost model. Several tests were conducted to assess the impact of lane width on crash rates. The RF model results showed an increase in crash rates as the lane width decreased. Using crash counts as the dependent variable (instead of the crash rate) led to a similar result. However, crash data with more detailed information are required to evaluate different contributing factors more comprehensively. For instance, information such as driver distraction, driver aggressiveness, and adverse weather conditions, if included in the crash data sets, would lead to a better understanding of all possible factors that may influence crash rates.